\documentclass[conference]{IEEEtran}
\IEEEoverridecommandlockouts
\usepackage{cite}
\usepackage{amsmath,amssymb,amsfonts}
\usepackage{algorithmic}
\usepackage{graphicx}
\usepackage{textcomp}
\usepackage{xcolor}

\def\BibTeX{{\rm B\kern-.05em{\sc i\kern-.025em b}\kern-.08em
    T\kern-.1667em\lower.7ex\hbox{E}\kern-.125emX}}
\begin{document}

\title{FusionSort: Enhanced Cluttered Waste
Segmentation with Advanced Decoding and
Comprehensive Modality Optimization
\thanks{The authors would like to thank MBZUAI for providing the computational resources used in this research. We also gratefully acknowledge the Multispectral Waste (MSWaste) dataset team for sharing their dataset.}
}

\author{\IEEEauthorblockN{1\textsuperscript{st} Muhammad Ali}
\IEEEauthorblockA{\textit{MBZUAI, AbuDhabi, UAE} \\
muhammad.ali@mbzuai.ac.ae} 
\and
\IEEEauthorblockN{2\textsuperscript{nd} Omar Ali AlSuwaidi}
\IEEEauthorblockA{\textit{MBZUAI, AbuDhabi, UAE} \\
omar.alsuwaidi@mbzuai.ac.ae}

}
\maketitle

\begin{abstract}
In the realm of waste management, automating the sorting process for non-biodegradable materials presents considerable challenges due to the complexity and variability of waste streams. To address these challenges, we introduce an enhanced neural architecture that builds upon an existing Encoder-Decoder structure to improve the accuracy and efficiency of waste sorting systems. Our model integrates several key innovations: a Comprehensive Attention Block within the decoder, which refines feature representations by combining convolutional and upsampling operations. In parallel, we utilize attention through the Mamba architecture, providing an additional performance boost. We also introduce a Data Fusion Block that fuses images with more than three channels. To achieve this, we apply PCA transformation to reduce the dimensionality while retaining the maximum variance and essential information across three dimensions, which are then used for further processing. We evaluated the model on RGB, hyperspectral, multispectral, and a combination of RGB and hyperspectral data. The results demonstrate that our approach outperforms existing methods by a significant margin.

\end{abstract}


\section{Introduction}
The rapid increase in global waste production, particularly from non-biodegradable materials such as plastics, underscores the urgent need for more efficient recycling processes \cite{plastic_waste}. Automated waste sorting systems play a pivotal role in improving recycling efficiency and ensuring the safety of workers in waste management. However, these systems face significant challenges due to the diverse and cluttered nature of waste streams \cite{icpram_enhaned}. Waste objects are often irregularly shaped, broken, dirty, or overlapping, and translucent materials further complicating the process. This variability requires advanced perception systems capable of capturing intricate details and handling the complex dynamics of waste sorting environments. 
\label{sec:intro}
\begin{figure}[htbp]
  \centering
  \includegraphics[width=0.48\textwidth]{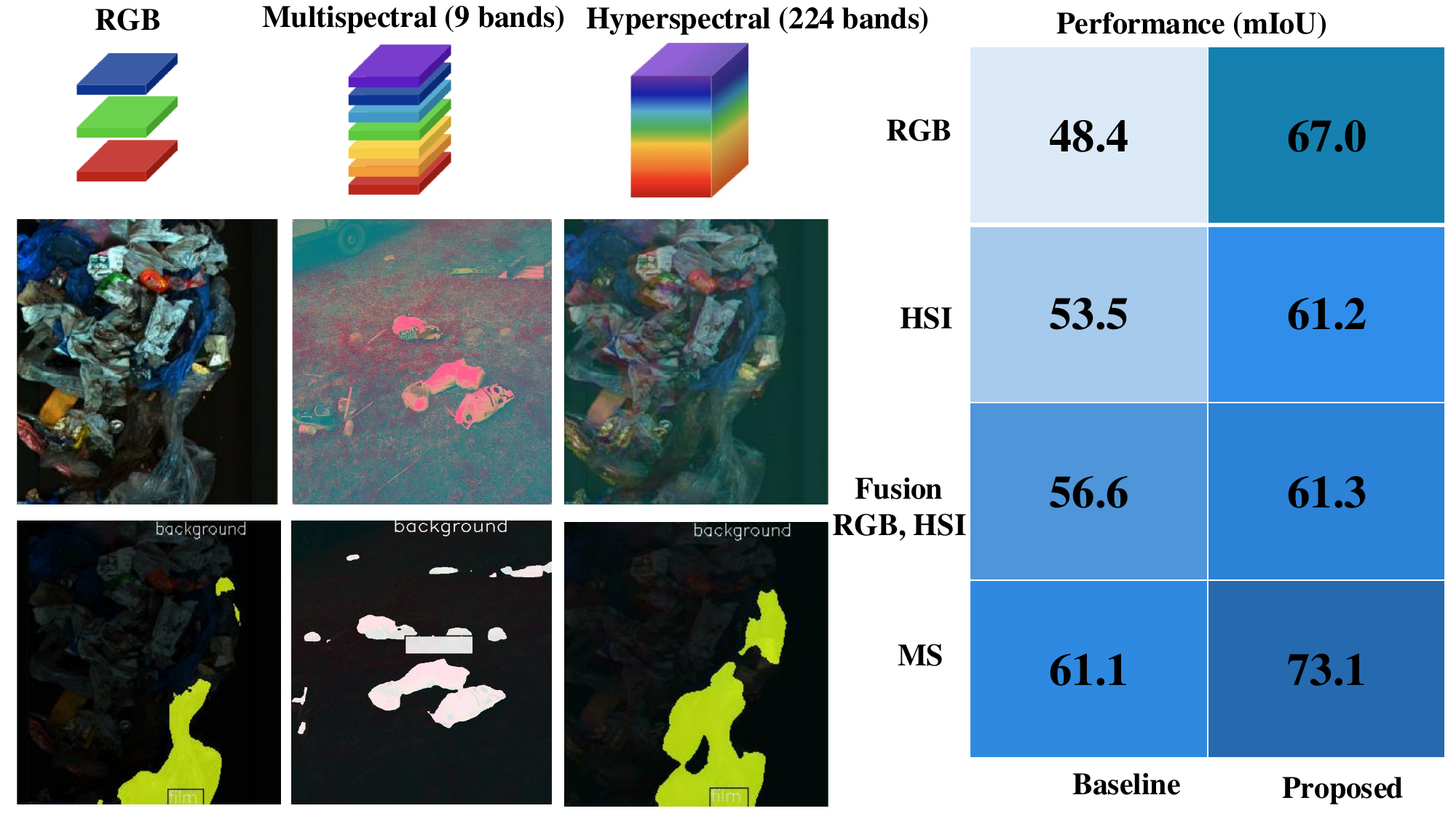} 
  \caption{Heatmap of segmentation performance (mIoU) for different data modalities like RGB, Multispectral(MS) \cite{zhu2024enhancing}, Hyperspectral(HSI) \cite{casao2024spectralwaste}, and their fusion. The heatmap illustrates significant performance improvements across all modalities.  Results show that the proposed method consistently outperforms baseline models, especially in the fused and Multispectral configurations. This demonstrates the advantage of leveraging spectral information and multimodal fusion for accurate waste object segmentation in cluttered and complex environments.}
  \label{fig:RGB_HSI_Fusion}
\end{figure}
Traditional sorting methods primarily rely on single-modality imaging techniques, such as RGB, which often fail to capture the subtle material differences essential for accurate classification. This limitation becomes particularly evident when objects in the waste stream share similar visual properties, making it difficult to differentiate them based solely on color or texture \cite{traditional_sorting}. Consequently, there is a growing need for sophisticated solutions that can leverage richer data to overcome these challenges and accurately identify materials under complex conditions.
As illustrated in Fig.\ref{fig:RGB_HSI_Fusion}, various color configurations and their respective band ranges are compared, including RGB, Multispectral, and Hyperspectral images, along with the corresponding input images and output masks, highlighting improvements across different modalities. Recent advances in multimodal imaging, particularly the combination of Hyperspectral Imaging (HSI) with RGB, offer a promising approach to address the limitations of single-modality systems. HSI captures detailed spectral information, allowing the differentiation of materials based on their unique spectral signatures, while RGB provides spatial information that helps define the shapes and boundaries of objects \cite{hsi_argiculture}. Together, these modalities provide a comprehensive representation of objects in the waste stream especially for some mateerails which are not fully covered by RGB, enhancing the potential for accurate segmentation even in cluttered environments. Although existing models have shown strong performance in general segmentation tasks, they often rely on single-modality data, limiting their ability to fully exploit the diverse information available from different imaging types. Our model, however, not only achieves state-of-the-art results across individual modalities, such as RGB and Hyperspectral, but also demonstrates excellent performance when these modalities are fused. Notably, our results indicate that the model's performance using only the RGB modality surpasses that of other methods, even in their fused versions, highlighting the robustness and efficiency of our approach. This further emphasizes the value of carefully designed architectures that can effectively utilize both individual and fused modalities for optimal waste segmentation.
FusionSort is a robust architecture for waste segmentation across diverse imaging modalities, including RGB, Hyperspectral Imaging (HSI), and their fusion. It enhances material differentiation by leveraging spatial and spectral information, improving segmentation accuracy in cluttered environments. The architecture features two key innovations: the Comprehensive Attention Block, which refines spatial and contextual relationships using attention mechanisms and Convolutional Coordinate Attention, and the Data Fusion Block, which applies Principal Component Analysis (PCA) to retain the most relevant spectral channels for optimal RGB-HSI integration. FusionSort sets a new benchmark in automated waste management, enabling precise and efficient material sorting to support sustainable recycling practices.

    Key contributions are as follows.
\begin{itemize}
    \item \textbf{FusionSort Architecture:} A novel neural network for fine waste segmentation across RGB, Multispectral, and Hyperspectral data, with enhanced accuracy through modality fusion.  

    \item \textbf{Comprehensive Attention Block:} An advanced attention mechanism integrating convolution, positional attention, and the Mamba Block to refine spatial and contextual relationships, improving segmentation in cluttered environments.  

    \item \textbf{Data Fusion:} A dedicated fusion block leveraging PCA for HSI dimensionality reduction, enhancing RGB-HSI integration and improving differentiation of visually similar waste materials.  

    \item \textbf{Evaluation on SpectralWaste and MultispectralWaste:} Rigorous assessment on SpectralWaste (224 HSI channels + RGB) and MultispectralWaste (9 spectral bands), demonstrating superior segmentation performance and generalizability.  
\end{itemize}

\section{Related Work}
This section reviews data sets and methods for multimodal data fusion, waste identification, and segmentation using attention mechanisms and CNNs.
\subsection{Waste Object Segmentation and Identification Datasets}  
Several datasets have been developed for waste management, focusing on object identification and segmentation. Stanford TrashNet \cite{trashnet} classifies single waste objects against plain backgrounds, suitable for basic tasks but lacking real-world complexity. TACO \cite{taco} improves on this with annotated images of waste in diverse settings, such as streets and beaches, supporting detailed litter analysis.
Specialized data sets such as floating waste (FloW) \cite{flow} offer multimodal data, including FloW-RI, which integrates millimeter wave radar with image data for better waste identification in aquatic environments. ZeroWaste \cite{zerowaste} and ZeroWaste-v2 \cite{zerowaste-v2}, developed for the VisDA2022 challenge, contain RGB images from industrial waste sorting, providing a solid base for object segmentation. Unlike these RGB-focused datasets, our FusioSort framework combines hyperspectral imaging (HSI) with RGB data, enhancing automated sorting systems by offering a richer source for waste identification and segmentation.

\subsection{Hyperspectral Imaging in Segmentation Tasks}
RGB images provide spatial clarity with basic color information, while multispectral images enhance spectral diversity by covering multiple wavelength bands. Hyperspectral images take this further, offering a continuous spectrum per pixel, ideal for precise identification of materials and subtle variations. Hyperspectral imaging (HSI) has been widely utilized in various fields due to its ability to capture detailed spectral information across a broad range of wavelengths. Traditional applications of HSI include water resource management \cite{zagolski1996forest,asner1998biophysical}, agriculture \cite{agri1,agri2} , medical \cite{med,med2} plant stress recognition \cite{kurata1996water} and recycling material classification \cite{bonifazi2006hyperspectral}, where additional spectral data provide insights not available through conventional RGB imaging \cite{hsi-overview}. 
In the context of waste management, HSI has been employed primarily for the identification of plastic materials, where per-pixel classification techniques are often used to label images based on their spectral properties \cite{bonifazi2006hyperspectral}.
 Unlike these methods focused on material. FusioSort further advances the application of HSI to object segmentation within industrial waste sorting, by ustilizing our architecture along with data fusion to improve segmentation accuracy.

\subsection{Multimodal Segmentation Techniques}

The fusion of multiple sensing modalities, such as combining RGB with HSI data, has been shown to significantly enhance the performance of classification tasks in various domains \cite{article,s19143071}. Similarly, in industrial applications, multimodal approaches have been applied to tasks such as the classification of building materials as well as  quality inspection of agricultural products \cite{inbook}. Despite these advancements, the application of multimodal segmentation techniques to real-world industrial waste sorting remains relatively unexplored. 
The fusion of multiple sensing modalities, such as combining RGB with hyperspectral imaging (HSI) data, has been shown to substantially enhance the performance of classification tasks across various domains, including medical imaging, remote sensing, and more recently, industrial applications \cite{article,s19143071}. Multimodal approaches have been successfully employed in areas like building material classification and agricultural product quality inspection, where the complementary strengths of different data types improve overall accuracy \cite{inbook}. However, the application of these advanced multimodal segmentation techniques to real-world industrial waste sorting remains largely underexplored.

In a recent effort to address this gap, Sara et al. introduced the SpectralWaste dataset, consisting  of synchronized RGB and HSI data collected from a real-world plastic waste sorting facility \cite{casao2024spectralwaste}. This dataset provides a critical resource for developing and evaluating multimodal approaches in waste management. Leveraging this dataset, we propose a novel architecture, FusioSort, which fuses RGB and HSI data to exploit the complementary benefits of these modalities. Our architecture is specifically designed to enhance object segmentation performance in cluttered and challenging environments, typical of waste sorting facilities.
FusioSort addresses the complexity of real-world waste sorting by utilizing the high spatial resolution of RGB images and the detailed spectral information from HSI, enabling more accurate and robust segmentation
\section{Method}
\label{sec:method}

\begin{figure*}[htb]
  \centering
   \includegraphics[width=\textwidth]{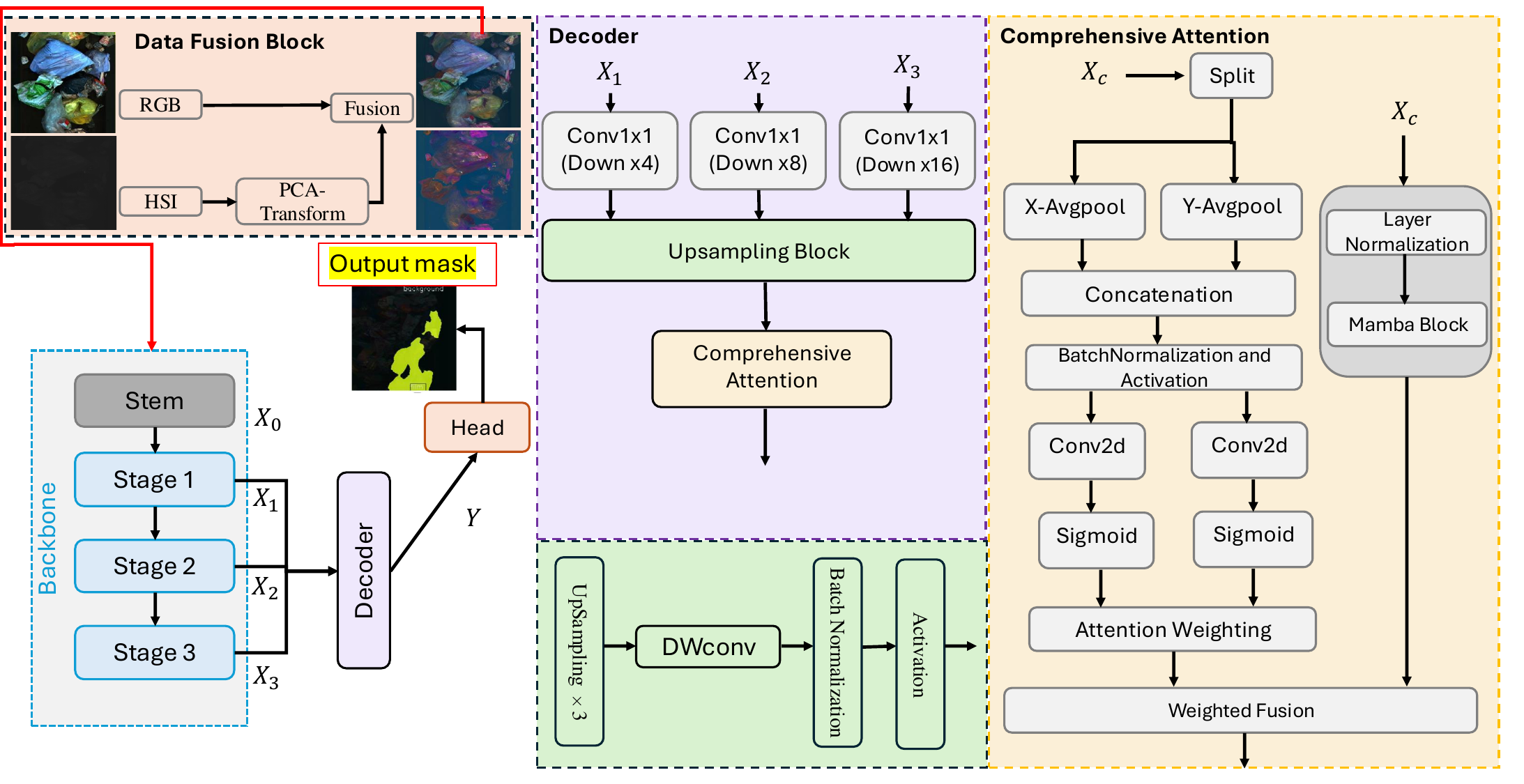}

   \caption{FusionSort Architecture for Multimodal Waste Sorting. The architecture includes  a hierarchical Backbone for feature extraction, and a Decoder with an Upsampling Block with integrated convolution to refine resolution and Comprehensive Attention Module. The Comprehensive Attention Module employs X- and Y-directional pooling, concatenation, and weighted fusion for precise feature refinement. Additionally, the Mamba Block applies attention mechanisms to further enhance the fused features, improving segmentation accuracy, and generating the final output mask.
}
   \label{fig:FusionSort}
\end{figure*}

In this section, we introduce FusionSort as given in  Fig. \ref{fig:FusionSort} which is built around an advanced Encoder-Decoder framework that integrates novel components to effectively leverage both hyperspectral imaging (HSI) and RGB data. The encoder is based on the existing transformer-based architecture utilized by Segformer \cite{neurips2021_segformer}, ensuring robust feature extraction, while the decoder incorporates innovative elements to enhance segmentation accuracy across diverse modalities.
The FusionSort architecture is composed of several key modules, each contributing to the model's robust performance as given in the Fig. \ref{fig:FusionSort}

\subsection{Data Fusion Block}


The Data Fusion Block is critical for synthesizing the multimodal inputs into a cohesive feature set. The process begins with the application of PCA to the hyperspectral data to get Hyper3,  effectively reducing its dimensionality while preserving the most informative spectral bands. This transformation is crucial for enabling real-time processing without compromising on the richness of the hyperspectral data.

The PCA-transformed HSI features are then resized to match the spatial resolution of the RGB features. These two data streams are fused, ensuring that the complementary information from both modalities is fully exploited. This fusion results in a feature set that captures both the detailed spectral information from the HSI and the spatial context from the RGB data, providing a robust foundation for subsequent segmentation tasks.

\subsection{Comprehensive Attention Block}
Comprehensive Attention Block: Integrated within the decoder stage, this block is designed to refine the feature maps produced by the encoder by focusing on spatial and contextual relationships in the data. 
This integrates two parallel attention mechanisms: Mamba attention and convolutional-based attention to further refined feature enhancement.
As illustrated in the Fig. \ref{fig:FusionSort}, the block begins by splitting the input feature map 
$X_c$
into separate components for processing. These components undergo X-Average Pooling and Y-Average Pooling, which extract spatial features along horizontal and vertical directions. The outputs are concatenated and processed through Batch Normalization and Activation layers, followed by two separate 2D convolutional layers. These operations compute spatial and contextual attention maps using sigmoid activations, ensuring the network selectively attends to the most relevant areas.
In parallel the input $X_c$ goes to Mamba block.


\begin{itemize}
    \item \textbf{Mamba Attetnion: }  We  utilized the Mamba attention mechanism to enhance feature representation efficiency and contextual understanding. Specifically, Mamba Attention was integrated into the decoder head to process the upsampled feature maps produced by the fusion block. This mechanism employs Layer Normalization followed by a state-space modeling (SSM) block and local convolutions. By using state expansion and local convolution width parameters, Mamba Attention effectively captures richer contextual relationships without incurring the computational overhead of traditional attention mechanisms. The incorporation of Mamba attention led to significant improvements in refining feature details, ultimately enhancing segmentation accuracy in complex and cluttered environments.
    

    \item \textbf{Channel Reduction and Weighted Attention Fusion:} The outputs of both parallel attention pathways are merged through channel reduction, reducing dimensionality while retaining critical information. This is followed by Weighted Attention Fusion, which combines the refined features into a cohesive representation, further enhancing the segmentation output and enabling precise classification even in complex and cluttered environments.

    \item \textbf{Final Processing:} The fused features undergo final refinement through batch normalization and activation functions, preparing them for the concluding stages of the decoder. This comprehensive approach ensures that the model maintains high precision in its segmentation predictions, effectively addressing the challenges of industrial waste sorting.
\end{itemize}

\subsection{Combined Dice and Cross Entropy Loss}

To optimize segmentation performance, FusionSort employs a hybrid loss function that synergizes the strengths of Dice Loss \cite{sudre2017generalised} and Cross Entropy Loss \cite{mao2023cross}.

The combined loss function is defined as follows:

\begin{equation}
\mathcal{L}_{\text{combined}} = \alpha \cdot \mathcal{L}_{\text{Dice}} + \beta \cdot \mathcal{L}_{\text{Cross Entropy}}
\end{equation}

where \(\mathcal{L}_{\text{Dice}}\) is the Dice Loss, \(\mathcal{L}_{\text{Cross Entropy}}\) is the Cross Entropy Loss, and \(\alpha\) and \(\beta\) are hyperparameters that control the trade-off between the two loss components. For our case we used\(\alpha\) value of 1 and \(\beta\) value of 1. 

\section{Experiments and Results}

\subsection{{Experimental Settings}}
We evaluate the performance of various models and configurations on the task of industrial waste segmentation. We benchmark our model on spectralwaste dataset and MultispectralWaste dataset.  
The proposed method is implemented in PyTorch using the MMSegmentation library \cite{mmseg2020}. Experiments were conducted on an NVIDIA RTX A6000 GPU. 
We optimized the model using the AdamW optimizer, starting with an initial learning rate of 9e-5, which was decayed using a polynomial scheduler. The model was trained for 60k iterations on the Spectral Waste dataset in three configurations: RGB, HSI, and RGB + HSI.  Images were resized to a 512 × 512 pixel. While for MSWaste model was traiined for 28k iterations.
To evaluate the performance of the proposed framework on cluttered scenes, we utilized SpectralWaste dataset \cite{spectralwaste} and MSWaste dataset.

\subsection{{Dataset}} SpectralWaste \cite{casao2024spectralwaste} is recently introduced dataset which has a cluttered background and translucent waste objects. It is collected from a waste processing facility and contains synchronized RGB and hyperspectral images. The dataset comprises of six object classes containing cardboard, film, basket, filaments, video tape, and trash bags. The dataset is challenging due to the presence of clutter and brightness variations. SpectralWaste dataset contains 514, 167, and 171 images in train, validation, and test sets respectively. 
The primary metric used for evaluation is the mean Intersection over Union (mIoU).

\subsection{{Results}} 

Our baseline models using Segformer, RGB-Baseline and HSI-Baseline, serve as foundational implementations that use only RGB and HSI(HyperSpectral Images) data, respectively. These baselines were designed to assess the effectiveness of conventional segmentation techniques without the integration of advanced architectural enhancements. The RGB-Baseline achieved a mIoU of 61.73\%, demonstrating solid performance, particularly in categories like Basket (76.83\%) and Filament (61.82\%).

Similarly, the HSI-Baseline, which focuses on hyperspectral data, attained a mIoU of 50.90\%. Although this baseline performed well in categories like Basket (72.67\%), the overall performance highlights the challenges of using hyperspectral data alone for complex waste segmentation tasks.\\
\subsubsection{Enhancements with FusionSort}
To improve upon the baseline results, we integrated our proposed modules, including the Comprehensive Attention Block and a combined Dice and Cross Entropy loss function. These enhancements were applied to both the RGB and HSI baselines.
Our proposed FusionSort architecture significantly enhances segmentation performance across individual RGB, HSI, and combined RGB+HSI datasets by incorporating the Comprehensive Attention Block and a combined Dice and Cross Entropy loss function as given in Table 1.

When applied to the RGB dataset, FusionSort demonstrated marked improvements over the RGB-Baseline, elevating the mean Intersection over Union (mIoU) from 61.73\% to 67.26\%. Notably, the model achieved better performance in the Cardboard category (75.52\%) and substantial gains in the Tape category (42.34\%) as well as in all other categories.
Similarly, FusionSort's impact on the HSI dataset where mIoU increased to 61.02\% compared to the 50.90\% achieved by the Hyper3-Baseline. In particular, the Cardboard category saw notable improvement (83.73\%), as did the Filament category (64.13\%).
Beyond individual modalities, the FusionSort architecture's ability to handle multimodal data was particularly effective. When applied to the fused RGB+HSI dataset, FusionSort improved the baseline mIoU from 55.83\% to 61.30\%, showing the value of combining modalities within a unified architecture. This multimodal configuration excelled in challenging categories such as Cardboard (86.12\%) and Bag (56.61\%).

The introduction of attention mechanisms, particularly the Comprehensive Attention Block, was key to these enhancements. By prioritizing and refining features from both RGB and HSI inputs, FusionSort effectively leveraged the strengths of each modality, resulting in a robust waste segmentation solution. This improvement demonstrates that the architecture not only enhances performance in individual modalities but also maximizes the potential of multimodal data fusion.
Overall, the results validate the effectiveness of the FusionSort architecture in delivering superior segmentation performance across RGB, HSI, and fused datasets. The gains across multiple categories and datasets underscore the importance of advanced attention mechanisms and multimodal fusion in tackling complex waste sorting challenges

\begin{table*}[htb]
\centering
\caption{Performance comparison of segmentation models and modalities on various waste categories using mIoU. The table shows results for RGB, hyperspectral (Hyper3), and multimodal fusion (RGB-Hyper3) across selected backbone architectures: MiniNet-v2, SegFormer-B0, CMX-B0, InternImage-T, and our proposed FusionSort. The chosen models represent a range of lightweight to advanced architectures relevant for segmentation tasks.}
\label{tab:SOTA_Hyper}
\scalebox{.9}{
\setlength{\tabcolsep}{4pt}
\begin{tabular}{l l c c c c c c c c}
\hline
Backbone & Modality & Film (\%)$\uparrow$ & Basket (\%)$\uparrow$ & Card. (\%)$\uparrow$ & Tape (\%)$\uparrow$ & Fil. (\%)$\uparrow$ & Bag (\%)$\uparrow$ & mIoU (\%)$\uparrow$ & Para (M) \\
\hline
\multicolumn{10}{c}{\textbf{RGB}} \\
\hline
MiniNet-v2 \cite{mininet2020} & RGB & 63.1 & 58.9 & 55.4 & 30.6 & 10.0 & 49.2 & 44.5 & 0.522 \\
SegFormer-B0 \cite{neurips2021_segformer} & RGB & 66.9 & 71.3 & 48.9 & 33.6 & 15.2 & 54.6 & 48.4 & 3.716 \\
\textbf{FusionSort} & \textbf{RGB} & \textbf{76.09} & \textbf{79.92} & \textbf{75.52} & \textbf{42.34} & \textbf{67.66} & \textbf{62.03} & \textbf{67.26} & \textbf{6.284} \\
\hline
\multicolumn{10}{c}{\textbf{Hyper3}} \\
\hline
MiniNet-v2 & Hyper3 & 58.8 & 61.9 & 69.4 & 30.2 & 23.0 & 50.5 & 49.0 & 0.522 \\
SegFormer-B0 & Hyper3 & 60.4 & 58.4 & \textbf{86.6} & 22.9 & 43.0 & 49.9 & 53.5 & 3.717 \\
\textbf{FusionSort} & \textbf{Hyper3} & \textbf{61.94} & \textbf{73.27} & 83.73 & \textbf{34.18} & \textbf{64.13} & \textbf{53.54} & \textbf{61.02} & \textbf{6.284} \\
\hline
\multicolumn{10}{c}{\textbf{RGB-Hyper3}} \\
\hline
MiniNet-v2 & RGB-H3 & 57.9 & 53.3 & 69.6 & 13.5 & 6.5 & 49.6 & 41.7 & 0.523 \\
SegFormer-B0 & RGB-H3 & 57.7 & 59.2 & 80.9 & 10.9 & 34.4 & 48.6 & 48.5 & 3.721 \\
InternImage-T \cite{wang2023internimage} & RGB-H3 & 51.99 & 45.38 & 85.8 & \textbf{71.1} & 44.39 & 18.5 & 40.77 & 10.647 \\
CMX-B0 \cite{zhang2023cmx} & RGB-H3 & \textbf{71.7} & 71.6 & 71.7 & 27.8 & 37.7 & \textbf{59.4 }& 56.6 & 11.193 \\
\textbf{FusionSort} & \textbf{RGB-H3} & 66.09 & \textbf{72.48} & \textbf{86.12} & 24.68 & \textbf{55.79} & 56.61 & \textbf{61.30} & \textbf{6.285} \\
\hline
\end{tabular}
}
\vspace{-1.0em}
\end{table*}

\subsection{Quantitative and Qualitative Analysis}
\subsubsection{SpectralWaste}
The performance analysis of our proposed FusionSort architecture shows improved performance across all modalities (RGB, HSI) and their fused configuration (RGB+HSI), establishing it as a more effective solution compared to state-of-the-art lightweight to moderate architectures such as MiniNet-v2, SegFormer-B0, InternImage-T and CMX-B0. The results detailed in the Table \ref{tab:SOTA_Hyper}  demonstrate how FusionSort consistently outperforms these competitors in a range of waste categories.

In the RGB modality, FusionSort achieves an mIoU of 67.26\%, significantly higher than MiniNet-v2 (44.5\%) and SegFormer-B0 (48.4\%). FusionSort excels in handling high-contrast objects like Cardboard (75.52\%) and Filament (67.66\%), demonstrating its capacity to leverage RGB data for effective segmentation. MiniNet-v2 and SegFormer-B0, on the other hand, struggle with intricate categories, particularly in segmenting fine details of Tape and Filament, highlighting the limitations of their feature extraction capabilities.

When evaluating the HSI modality, FusionSort again performs well with an mIoU of 61.02\%, outperforming SegFormer-B0 53.5\% and CMX-B0 (56.6\%). The HSI configuration of Fusion Sort is particularly effective in distinguishing subtle material variations, such as those in Filament (64.13\%) and Bag (53.54\%), where spectral data plays a crucial role. This demonstrates FusionSort's enhanced ability to exploit hyperspectral information compared to other models, such as InternImage-T, which achieves only 40.77\% mIoU and struggles significantly in texture-intensive categories.

The fused RGB+HSI configuration of FusionSort achieves the highest overall performance with an mIoU of 61.30\%, demonstrating balanced and robust segmentation across diverse waste categories. This configuration leverages the complementary strengths of both RGB’s spatial detail and HSI’s spectral information, resulting in the most precise segmentation boundaries, particularly in challenging categories such as Cardboard (86.12\%) and Filament (55.79\%). In contrast, CMX-B0, despite its advanced architecture, achieves a lower mIoU of 56.6\% and faces challenges in consistently handling complex material features across categories

\begin{figure*}[htb]
  \centering
  \includegraphics[trim=0 0 0 0, clip, width=\textwidth]{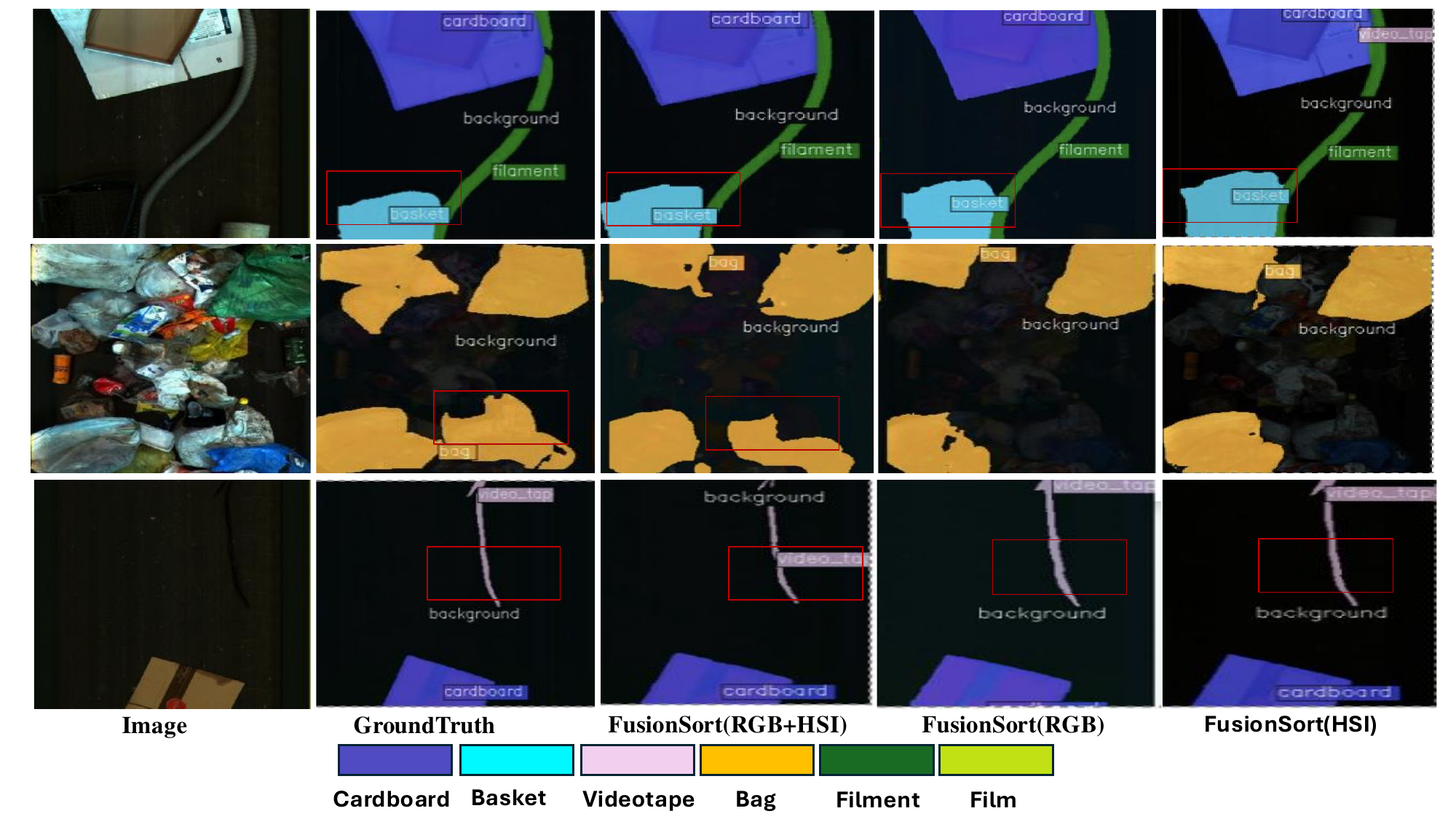}
  \caption{Comparison of segmentation results across three configurations of the proposed FusionSort architecture: FusionSort (RGB + HSI), FusionSort (RGB), and FusionSort (HSI). The results demonstrate the model's ability to segment and classify waste categories, including Cardboard, Basket, Videotape, Bag, Filament, and Film. The FusionSort (RGB + HSI) configuration outperforms the single-modality setups by effectively leveraging the complementary strengths of RGB and hyperspectral data, resulting in more accurate segmentation and classification. FusionSort (RGB) performs well in capturing spatial details but lacks spectral precision, while FusionSort (HSI) is effective in highlighting material properties but struggles with spatial consistency. This comparison underscores the importance of multimodal data fusion for robust and accurate waste segmentation.}
  \label{fig:Eval_Fusionsort}
\end{figure*}

Fig.\ref{fig:Eval_Fusionsort} illustrates the robust performance of FusionSort across all configurations, highlighting the architecture’s adaptability and effectiveness in various data settings. In the RGB configuration, FusionSort accurately segments high-contrast objects such as Videotape, Cardboard, and Filament, demonstrating clear boundaries that reflect the model’s strength in color-based differentiation. Although some challenges remain with texture-dependent materials like Bag, FusionSort’s consistent boundary precision underscores its capability to handle complex scenes using RGB data alone.

The HSI configuration further showcases FusionSort’s ability to excel in segmenting nuanced materials, such as Bag and Filament, by leveraging hyperspectral data. The Fig. confirms how FusionSort can accurately highlight subtle spectral variations, which are essential for differentiating complex materials that RGB struggles to capture. Even with spectral data alone, the model maintains strong segmentation performance, emphasizing its capacity to adapt and optimize across diverse data inputs.

The fused RGB+HSI configuration combines the best of both worlds, achieving the most balanced and precise segmentation boundaries, as seen in challenging categories like Cardboard and Basket. However, the Fig. clearly demonstrates that FusionSort excels not only in multimodal fusion but across all configurations, setting it apart from other architectures. This consistent performance across modalities confirms that FusionSort is highly effective at leveraging the strengths of each data type, making it a standout solution for complex waste sorting scenarios.

Overall, the qualitative insights align closely with the quantitative performance metrics, confirming that FusionSort consistently delivers better segmentation results across various categories and modalities. The combination of advanced attention mechanisms and multimodal fusion in FusionSort is key to its success in addressing complex waste sorting challenges.
The Comprehensive Attention Block helps  highlighting relevant features and suppressing less significant ones, allowing the model to focus on critical areas. This is particularly valuable in waste sorting, where cluttered scenes and visually similar objects challenge segmentation accuracy. It further  preserves essential structural and spectral information, refining features through convolutional integration and weighted attention fusion. This maximizes the use of spatial and spectral data, boosting performance across RGB, HSI, and combined configurations.

\begin{figure*}[htb]
  \centering
  \includegraphics[width=\textwidth]{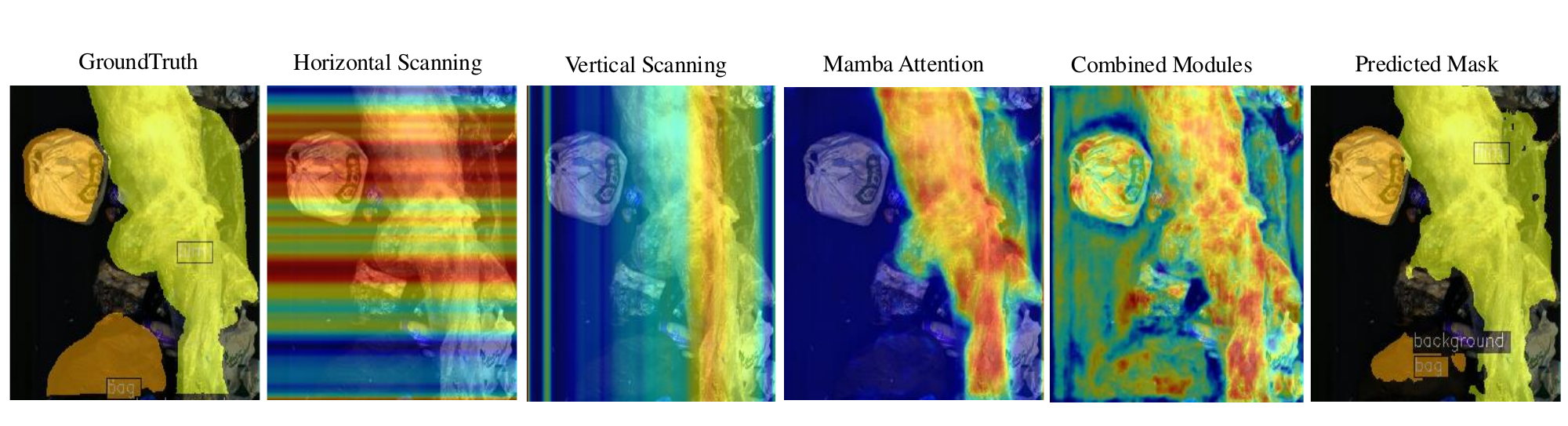}
  \caption{This Fig. presents a comprehensive analysis of the Fusionsort decoder's feature maps. It represents feature maps obtained from comprehensive attention block mentioned as horizontal scanning and vertical scanning, followed by Mamba attention block and weighted attention fusion output masks. It also includes the final segmentation prediction generated by the model and the corresponding ground truth for comparison. }
  \label{fig:features}
\end{figure*}

The feature visualizations  at the decoder level given in Fig. \ref{fig:features} demonstrate a stepwise enhancement of information representation throughout the model’s decoding process. Initially, horizontal and vertical scanning are used to capture directional features, contributing to an enriched spatial understanding. This helps emphasize distinct patterns in the waste materials. The subsequent attention mechanisms, including comprehensive attention, enable the model to focus on important regions effectively, refining the final feature maps to closely align with the ground truth. These progressive visualizations highlight the gradual accumulation of contextual details, ultimately improving segmentation accuracy across challenging and cluttered environments.

\subsubsection{MSWASTE}
\begin{figure*}[htb]
  \centering
  \includegraphics[scale=.3]{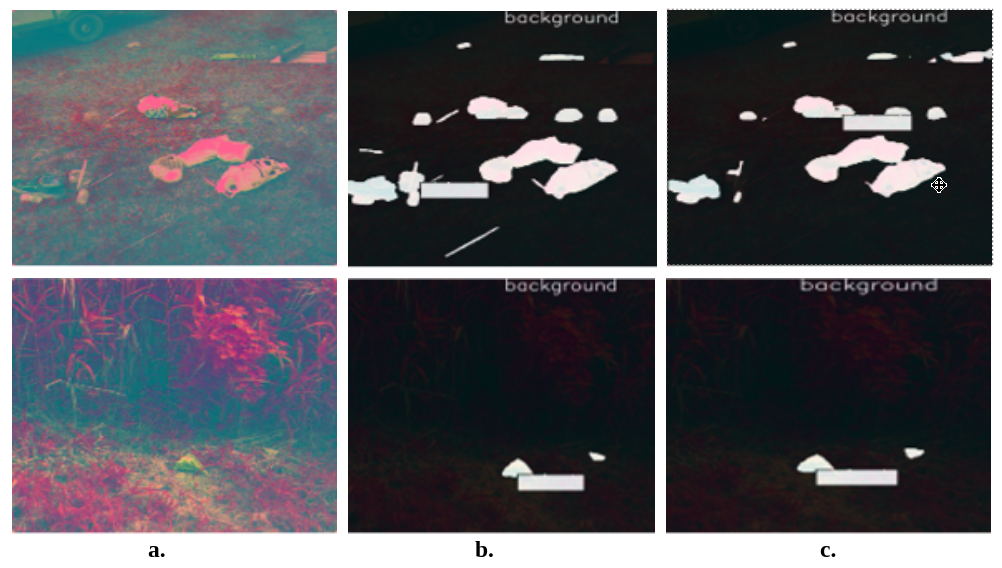}
  \caption{Results on the Multispectral Waste Dataset. The Fig. shows the Original Image (a), the Ground Truth segmentation map (b), and the Prediction by the model (c), illustrating the model’s ability to accurately segment waste materials in multispectral imagery.}
  \label{fig:MS}
\end{figure*}

The WasteMS dataset contains 117 multispectral images, each with nine channels and a resolution of 682×682 pixels, accompanied by semantic segmentation annotations. For effective deep learning training and evaluation, the dataset was divided into training, validation, and test sets, following an approximate ratio of 7:1:2. This standard split facilitates robust training and reliable assessment of semantic segmentation models.
The results for the MultispectralWaste dataset are summarized in Table \ref{tab:model_performance}. Our approach demonstrates a significant performance improvement, achieving an accuracy increase of up to 73\% compared to existing models. Notably, our model achieves this enhanced performance with fewer parameters, showcasing its efficiency. However, the computational cost, as measured by FLOPs, is higher due to the added complexity in feature processing, which contributes to the observed accuracy gains.
Fig.   \ref{fig:MS} presents the visualization results of our model's predictions on the Multispectral Waste Dataset. Since the actual images are 9-channel multispectral images, they have been pseudocolored for visualization purposes. Each row depicts an input image (a), the corresponding ground truth segmentation map (b), and the predicted segmentation map (c).

The top row presents a complex scene featuring scattered and multi-scaled waste along a lakeside. In this challenging environment, our model effectively segments the waste materials, closely aligning with the provided ground truth annotations. The accurate delineation of diverse waste types, despite their scattered distribution and varying scales, demonstrates the model's capability to manage intricate spatial relationships within natural environments.

Similarly, the second row illustrates another lakeside waste image, characterized by the presence of small-scale waste materials. In this scenario, our model successfully segments the waste, highlighting its proficiency in dealing with scale variations and complex settings. The model accurately identifies and segments waste items despite their smaller size and challenging background, further demonstrating its robustness.

These results collectively underscore the strength of our model in processing and segmenting multispectral imagery, especially in natural environments where waste exhibits diverse scales, scattering, and intricate context. The ability to maintain high segmentation accuracy across these complex scenes reflects the model's reliability and versatility for real-world waste management applications.
\begin{table}[htb]
\centering
\caption{Performance comparison of segmentation models with pretrained weights using different encoder-decoder combinations, evaluated with IoU, parameters}
\label{tab:model_performance}
\scalebox{0.85}{
\setlength{\tabcolsep}{4pt}
\begin{tabular}{l l c c c}
\hline

\textbf{Encoder} & \textbf{Decoder} & \textbf{IoU(\%)} & \textbf{\#Params (M)}   \\ \hline
ResNet50 & FCN & 43.62 & 47.12  \\
ResNet50 & PSPNet & 45.51 & 46.60  \\
ResNet50 & DeepLabV3+ & 55.03 & 41.22  \\
ResNet50 & UperNet & 61.77 & 64.04  \\
ConvNeXt-T & UperNet & 59.63 & 59.25  \\
Swin-T & UperNet & 57.93 & 58.95 \\
{MixVisionTrans.} & \textbf{FusionSort} & \textbf{73.12} & \textbf{6.696}  \\
\hline
\end{tabular}
}
\vspace{-1.0em}
\end{table}

\subsubsection{Ablation Study}
The ablation study conducted on the MultispectralWaste dataset evaluates the contribution of individual components in the proposed architecture, as shown in Table \ref{tab:model_performance}
. Starting with the baseline configuration, which achieves a mean Intersection over Union (mIoU) of 62.34\% and Pixel Accuracy of 91.23\%, the introduction of Mamba attention shows good improvement.

The Comprehensive Attention, when added, provides a more focused feature refinement, leading to a marked increase in mIoU and Pixel Accuracy. This component significantly enhances the model's capacity to differentiate overlapping waste categories, effectively highlighting the benefits of advanced contextual aggregation for improved feature representation. Weighted Fusion further enhances the results, demonstrating the effectiveness of integrating features.

Finally, the inclusion of all components Mamba attention, Comprehensive Attention, and Weighted Fusion demonstrates that these modules are complementary. The highest mIoU and Pixel Accuracy scores, 70.51\% and 93.45\% respectively, reflect how the combined architecture capitalizes on each component's strength, yielding a more holistic and accurate segmentation outcome. These findings emphasize the integrated approach's ability to effectively tackle the challenges associated with segmenting waste in diverse and unstructured environments.

While FusionSort demonstrates strong performance across RGB, HSI, and fused modalities, certain limitations remain. Specifically, the model exhibits lower performance on visually ambiguous or thin materials such as "Tape," where low inter-class contrast or limited spectral differentiation may hinder precise segmentation. This is likely due to subtle reflectance differences between similar plastic types that even fused modalities cannot easily resolve. Additionally, in real-world scenarios, factors such as occlusion, motion blur in conveyor-based systems, or insufficient lighting may further reduce segmentation accuracy. Moreover, the high cost, bulkiness, and calibration requirements of hyperspectral cameras may limit the immediate deployment of HSI-based models in industrial waste sorting systems. Future work will explore model optimization for real-time inference and robustness under practical constraints, including reduced sensor modalities or compressed spectral bands.

\begin{table*}[htb]
\centering
\caption{Ablation study results evaluating the impact of individual and combined modules in the proposed segmentation architecture. The configurations include the Baseline model and incremental additions of the Mamba Block, Comprehensive Attention, and Weighted Fusion modules. The performance is measured using mean Intersection over Union (mIoU) and Pixel Accuracy, both expressed as percentages. A tick (\checkmark) indicates that the corresponding module is applied in the configuration, while a dash (–) indicates it is not applied.}
\label{tab:ablation_study}
\scalebox{.8}{
\setlength{\tabcolsep}{6pt}
\begin{tabular}{l c c c c c}
\hline
\textbf{Configuration} & \textbf{Comprehensive Attention} & \textbf{Mamba} & \textbf{Weighted Fusion} & \textbf{mIoU (\%)} & \textbf{Pixel Accuracy (\%)} \\
\hline
Baseline                    & – & – & – & 62.34 & 91.23 \\
Mamba                       & – & \checkmark    & – & 67.56 & 92.06 \\
Comprehensive Attention     & \checkmark    & – & – & 69.04 & 92.62 \\
Weighted Fusion             & – & – & \checkmark   & 69.90 & 92.86 \\
All Modules                 & \checkmark   & \checkmark    & \checkmark    & 70.51 & 93.45 \\
\hline
\end{tabular}
}
\vspace{-1.0em}
\end{table*}

\section{Conclusion}

Our work introduces an innovative approach to waste segmentation through the fusion of RGB and hyperspectral imaging (HSI), supported by a targeted architectural design aimed at improving segmentation accuracy. By effectively integrating RGB and HSI data, the proposed method addresses the inherent challenges of waste streams, particularly for non-biodegradable materials, and achieves enhanced material differentiation. Key contributions, including the Comprehensive Attention Block for precise feature refinement and the Data Fusion Block for seamless multimodal integration, have proven instrumental in driving these advancements.
The model was rigorously evaluated on both the SpectralWaste and MultispectralWaste datasets, consistently outperforming traditional segmentation techniques. 
\bibliographystyle{IEEEtran}
\bibliography{main.bib}

\end{document}